%% file: learning_recon.tex
\documentclass{mrmarxiv}
\setboolean{bDraft}{true} 
\setboolean{bComments}{true}

\input{def_math}
\input{def_figures}

\graphicspath{{./}{./figures/}}

\title{Learning a Variational Network\\ for Reconstruction of
	   Accelerated MRI Data}

\runningtitle{Learning a Variational Network for Reconstruction of Accelerated MRI Data}

\author{Kerstin~Hammernik$^{1*}$, Teresa~Klatzer$^1$, Erich~Kobler$^1$,\\ Michael~P~Recht$^{2,3}$, Daniel~K~Sodickson$^{2,3}$,\\
    Thomas~Pock$^{1,4}$ and Florian~Knoll$^{2,3}$}

\affiliation{$^1$ Institute of Computer Graphics and Vision,\\Graz University of
	Technology, Graz, Austria
	
	\affilationspace $^2$ Center for Biomedical Imaging, Department of Radiology,\\ NYU School of Medicine, New
	York, NY, United States
	
	\affilationspace $^3$ Center for Advanced
	Imaging Innovation and Research (CAI$^2$R),\\ NYU School of Medicine, New
	York, NY, United States
	
	\affilationspace $^4$ Center for Vision, Automation \& Control,\\ AIT Austrian
	Institute of Technology GmbH, Vienna, Austria}

\correspondence{
	$^*$Correspondence to: Kerstin Hammernik\\
	Institute of Computer Graphics and Vision\\
	Graz University of Technology\\
	Inffeldgasse 16/II, 8010 Graz, Austria\\
	Phone: +43 316 873-5057, Fax: +43 316 873-5050\\
	E-mail: hammernik@icg.tugraz.at \\
	
	\noindent Preliminary data for this article were presented at the 24th Annual Meeting of ISMRM, Singapore, 2016.}

\grant{FWF START Project BIVISION, No. Y729; ERC starting grant ”HOMOVIS”, No. 640156.; NIH P41 EB017183; NIH R01 EB000447.}
\numWordsAbstract{193}
\numWordsBody{Approx. 5000}
\numFigures{10}
\numCitations{54}

\keywords{Variational Network; Deep Learning; Accelerated MRI; Parallel Imaging; Compressed Sensing; Image Reconstruction}

\begin{document}
	
	{
	\linespread{1.3}\maketitle
	}
	
	\begin{abstract}
		\paragraph{Purpose:} To allow fast and high-quality reconstruction of clinical accelerated multi-coil MR data by learning a variational network that combines the mathematical structure of variational models with deep learning.
		
		\paragraph{Theory and Methods:} Generalized compressed sensing reconstruction formulated as a variational model is embedded in an unrolled gradient descent scheme. All parameters of this formulation, including the prior model defined by filter kernels and activation functions as well as the data term weights, are learned during an offline training procedure. The learned model can then be applied online to previously unseen data.
		
		\paragraph{Results:} The variational network approach is evaluated on a clinical knee imaging protocol. The variational network reconstructions outperform standard reconstruction algorithms in terms of image quality and residual artifacts for all tested acceleration factors and sampling patterns.
					
		\paragraph{Conclusion:}
		Variational network reconstructions preserve the natural appearance of MR images as well as pathologies that were not included in the training data set. Due to its high computational performance, i.e.,~reconstruction time of 193 ms on a single graphics card, and the omission of parameter tuning once the network is trained, this new approach to image reconstruction can easily be integrated into clinical workflow.
		
	\end{abstract}
	
	\section{Introduction} 
	Imitating human learning with deep learning~\cite{LeCun2015,Goodfellow2016} has become an enormously important area of research and development, with a high potential for far-reaching application, including in the domain of Computer Vision. Taking encouragement from early successes in image classification tasks~\cite{Krizhevsky2012}, recent advances also address semantic labeling~\cite{Chen2015a}, optical flow~\cite{Dosovitskiy2015} and image restoration~\cite{Chen2015}. In medical imaging, deep learning has also been applied to areas like segmentation~\cite{Zhang2015,Moeskops2016}, q-space image processing~\cite{Golkov2016}, and skull stripping~\cite{Kleesiek2016}. However, in these applications, deep learning was seen as a tool for image processing and interpretation. The goal of the current work is to demonstrate that the concept of learning can also be used at the earlier stage of image formation. In particular, we focus on image reconstruction for accelerated MRI, which is commonly accomplished with frameworks like Parallel Imaging (PI)~\cite{Sodickson1997,Pruessmann1999,Griswold2002} or Compressed Sensing (CS)~\cite{Candes2006,Donoho2006,Lustig2007}. CS in particular relies on three conditions to obtain images from k-space data sampled below the Nyquist rate~\cite{Nyquist1928,Shannon1949}.
		
	The first CS condition requires a data acquisition protocol for	undersampling such that artifacts become incoherent in a certain transform domain~\cite{Candes2006,Donoho2006}. In MRI, we usually achieve incoherence by random~\cite{Lustig2007} or non-Cartesian sampling trajectories~\cite{Block2007}. The second requirement for CS is that the image to be reconstructed must have a sparse representation in a certain transform domain. Common choices are the Wavelet transform~\cite{Lustig2007,Daubechies1992} or Total Variation (TV)~\cite{Block2007,Rudin1992,Knoll2010a,Knoll2012}. In these transform domains, the $l_1$ norm is commonly applied to obtain approximate sparsity. The third CS condition requires a non-linear reconstruction algorithm that balances sparsity in the transform domain against consistency with the acquired undersampled k-space data.
	
	Despite the high promise of CS approaches, most routine clinical MRI examinations are still based on Cartesian
	sequences. Especially in the case of 2D sequences, it can be challenging to fulfill the criteria for incoherence required by CS~\cite{Hollingsworth2015}.
	One other obstacle to incorporation of CS into some routine clinical routine examinations
	is the fact that the sparsifying transforms employed in CS applications to date may be too simple to capture the complex
	image content associated with biological tissues. This can lead to reconstructions that appear blocky and unnatural, which reduces acceptance by clinical radiologists.
	A further drawback, not only for CS but for advanced image acquisition and reconstruction methods in general, is the long image reconstruction time typically required for iterative solution of non-linear optimization problems.
	A final challenge concerns the selection and tuning of hyper-parameters for CS approaches. A poor choice of hyper-parameters leads either to over-regularization, i.e.,~excessively smooth
	or unnatural-looking images, or else to images that still show residual undersampling
	artifacts. The goal of our current work is to demonstrate that, using learning approaches, we can achieve accelerated and high-quality MR image reconstructions from undersampled data which do not fulfill the usual CS conditions.

	With current iterative image reconstruction approaches, we treat every single exam and resulting image reconstruction task as a new
	optimization problem. We do not use information about the expected appearance of the anatomy, or the known structure of undersampling artifacts, explicitly in these optimization problems, which stands in stark contrast to how human radiologists read images. Radiologists are trained throughout their careers to look for certain reproducible patterns, and they obtain remarkable skills to ``read through'' known image artifacts~\cite{Hollingsworth2015}. Essentially, they rely on prior
	knowledge of a large number of previous cases, and they develop these skills by reading thousands of cases over the course of their careers. Translating this learning experience to deep learning allows us to shift the key effort of optimization from the online reconstruction stage to an up-front offline training task. In other words, rather than solving an inverse problem to compute, for each new data set, a suitable transform between raw data and images, we propose to \emph{learn} the key parameters of that inverse transform in advance, so that it can be applied to all new data as a simple flow-through operation.
	
	In this work, we introduce an efficient trainable formulation for accelerated PI-based MRI reconstruction that we term a \emph{variational network} (VN). The VN embeds a generalized CS concept, formulated as a variational model, within a deep learning approach.
	Our VN is designed to learn a complete reconstruction procedure for complex-valued multi-channel MR data, including all free
	parameters which would otherwise have to be set empirically.
	We train the VN on a complete clinical protocol for musculoskeletal imaging, evaluating performance for different
	acceleration factors, and for both regular and pseudo-random Cartesian 2D sampling.
	Using clinical patient data, we investigate the capability of the VN approach to preserve unique pathologies that are not included in the training data set.
	
	\vspace*{1\baselineskip} \section{Theory}

	\subsection{From Linear Reconstruction to a Variational Network}
	In MRI reconstruction, we naturally deal with complex numbers. Here, we introduce a mapping to real-valued numbers that we will use throughout our manuscript. We define complex images $\tilde{\vu}$ of size $n_{\textsc{x}}\times n_{\textsc{y}}=N$ as 	equivalent real images $\vu$ as follows:
	\begin{align*}
	\tilde{\vu} = \vu _\reu + \iu \vu _\imu \in \C^N \Leftrightarrow \vu=(\vu _\reu,\vu _\imu)\in\R^{2N}.
	\end{align*} 
	
	We consider the ill-posed linear inverse problem of finding a reconstructed image $\vu\in \R^{2N}$
	that satisfies the following system of equations
	\begin{align}\label{eq:system}
		A\vu = \hat{\vf},
	\end{align}
	where $\hat{\vf}\in\R^{2NQ}$ is the given undersampled k-space data, where missing data are padded by zeros. The linear forward sampling operator $A$ implements point-wise multiplications with $Q$ coil sensitivity maps, Fourier transforms, and undersampling according to a selected sampling pattern. Originally, the operator $A$ is defined by the mapping $\C^N \mapsto \C^{NQ}$, but embedding it in our real-valued problem changes the mapping to $\R^{2N} \mapsto \R^{2NQ}$.
	Since the system in~\eqref{eq:system} is ill-posed, we cannot solve for $\vu$ explicitly. Therefore, a natural idea is to compute $\vu$ by minimizing the least squares error
	\begin{align}
		\min_\vu \frac{1}{2} \norm[2]{A\vu- \hat{\vf}}^2.\label{eq:ls}
	\end{align}
	In practice we do not have access to the true $\hat{\vf}$ but only to a noisy variant $\vf$ satisfying
	\begin{align*}
		\norm[2]{\hat{\vf}-\vf} \leq \delta
	\end{align*}
	where $\delta$ is the noise level.
	The idea is to perform a gradient descent on the least squares problem~\eqref{eq:ls} that leads to an iterative algorithm, which is known as the Landweber method~\cite{Landweber1951}.
	It is given by choosing some initial $\vu^0$ and performing the iterations with step sizes $\alpha^t$
	\begin{align}
		\vu^{t+1} = \vu^t - \alpha^t A^*(A\vu^t - \vf), \quad t \geq 0
	    \label{eq:landweber}
	\end{align}
	where $A^*$ is the adjoint linear sampling operator.
	To prevent over-fitting to the noisy data $\vf$,
	it is beneficial to stop the Landweber iterative algorithm early~\cite{Hanke1995}, i.e.,~after a finite number of iterations $T$.
	
	Instead of early stopping, we can also extend the least squares problem by an additional regularization term $\calR(\vu)$ to prevent over-fitting.
	The associated (variational) minimization problem is given by
	\begin{align*}
		\min_\vu \left\lbrace \mathcal{R}(\vu) + \frac{\lambda}{2} \norm[2]{A\vu-\vf}^2\right\rbrace. 
	\end{align*}
	The minimizer of the regularized problem depends on the trade-off between the regularization term and the least squares data fidelity term controlled by $\lambda > 0$. One of the most influential regularization terms in the context of images is
	the TV semi-norm~\cite{Rudin1992}, which is defined as
	\begin{align*}
	\mathcal{R}(\vu) = \norm[2,1]{(D\vu_\reu,D\vu_\imu)} = \sum\limits_{l=1}^N\sqrt{\abs{D \vu_\reu}^2_{l,1} + \abs {D \vu_\imu}^2_{l,1} + \abs{D \vu_\reu}^2_{l,2} + \abs{D \vu_\imu}^2_{l,2}}
	\end{align*}
	where $D:\R^{N} \mapsto \R^{N\times 2}$ is a finite differences approximation of the image gradient, see for example~\cite{Chambolle2016}. The main advantage of TV is that it allows for sharp discontinuities (edges) in the solution while being a convex functional enabling efficient and global optimization. From a sparsity point of view, TV induces sparsity in the image edges and hence, favors piecewise constant solutions. However, it is also clear that the piecewise-constant approximation is not a suitable criterion to describe the complex structure of MR images and a more general regularizer is needed.
	
	A generalization of the TV is the Fields of Experts
	model~\cite{Roth2009}
	\begin{align}
		\mathcal{R}(\vu) = \sum_{i=1}^{N_k}\dotp{\Phi_i(K_i \vu)}{\mat{1}}.\label{eq:foe}
	\end{align}
	Here, the regularization term is extended to $N_k$ terms and $\mat{1}$ denotes a vector of ones.
	The linear operator $K = \left(K_{\reu}, K_{\imu}\right):\R^{2N} \mapsto \R^N$ models convolutions with filter kernels $k \in \R^{s\times s\times 2}$ of size $s$, which is expressed as
	\begin{align*}
	K\vu=K_{\reu} \vu_{\reu} + K_{\imu} \vu_{\imu},\quad \vu \in \R^{2N} \Leftrightarrow u\ast k = u_{\reu}\ast k_{\reu} + u_{\imu}\ast k_{\imu},\quad u\in\R^{n_{\textsc{x}}\times n_{\textsc{y}}\times 2}.
	\end{align*}
	The non-linear potential functions $\Phi(\vz) = \left(\phi(z_1),...,\phi(z_N) \right)^\top : \R^N \mapsto \R^N$ are composed by scalar functions $\phi$.
	In the Fields of Experts model~\cite{Roth2009}, both convolution kernels and parametrization of the non-linear potential functions, such as student-t functions, are learned from data.
	 
	Plugging the Fields of Experts model~\eqref{eq:foe} into the Landweber iterative algorithm~\eqref{eq:landweber} yields
	\begin{align}
	 		\vu^{t+1} = \vu^t - \alpha^t\left(\sum_{i=1}^{N_k}(K_i)^\top \Phi_i^{\prime}(K_i \vu^t) + \lambda A^*(A\vu^t - \vf)\right)\label{eq:reg_landweber}
	\end{align}
	where
	$\Phi^{\prime}_i(\vz)=\diag{\phi^{\prime}_i(z_1),...,\phi^{\prime}_i(z_N)}$ are the activation functions defined by the first derivative of potential functions $\Phi_i$. Observe that the application of the tranpose operation $(K_i)^\top$ can be implemented as a convolution with filter kernels $k_i$ rotated by 180\degree.
	Chen et al.~\cite{Chen2015} introduce a trainable reaction-diffusion approach that performs early stopping on the gradient scheme~\eqref{eq:reg_landweber} and allows the parameters, i.e.,~filters, activation functions and data term weights, to vary in every gradient descent step $t$. All parameters of the approach are learned from data. This approach has been successfully applied to a number of image processing tasks including image denoising~\cite{Chen2015}, JPEG deblocking~\cite{Chen2015}, demosaicing~\cite{Klatzer2016} and image inpainting~\cite{Yu2015}. For MRI reconstruction, we rewrite the trainable gradient descent scheme with time-varying parameters $K_i^t,\,\Phi_i^{t\prime},\,\lambda^t$ as
	\begin{align}
		\vu^{t+1} = \vu^t - \sum_{i=1}^{N_k}(K_i^t)^\top \Phi_i^{t\prime}(K_i^t \vu^t) - \lambda^t A^*(A\vu^t - \vf), \quad 0 \leq t \leq T-1.
		\label{eq:vn}
    \end{align}
 Additionally, we omit the step size $\alpha^t$ in~\eqref{eq:reg_landweber} because it is implicitly contained in the activation functions and data term weights.
	
	By unfolding the iterations of~\eqref{eq:vn}, we obtain the \emph{variational network} (VN) structure as depicted in Figure~\ref{fig:figure1-network}. Essentially, one iteration of an iterative reconstruction can be related to one step in the network. In our VN approach, we directly use the measured raw data as input. Coil sensitivity maps are pre-computed from the fully sampled k-space center. The measured raw data and sensitivity maps, together with the zero filled initializations, are fed into the VN as illustrated in Figure~\ref{fig:figure1-network}a. The sensitivity maps are used in the operators $A,A^*$, which perform sensitivity-weighted image combination and can also implement other processing steps such as the removal of readout oversampling. While both raw data and operators $A,A^*$ are required in every iteration of the VN to implement the gradient of the data term, the gradient of the regularization is only applied in the image domain (see Figure~\ref{fig:figure1-network}b).
	
	\insertmrmfig{\figI}
	
	\vspace*{1\baselineskip} \section{Methods}
	\subsection{Variational Network Parameters}
	The VN defined by~\eqref{eq:vn} and illustrated in Figure~\ref{fig:figure1-network}b contains a number of parameters: Filter kernels $k_i^t$, activation functions $\Phi_i^{t\prime}$, and data term weights 
	$\lambda^t$. We first consider the filter kernels which requires us to introduce a vectorized version $\vk_i^t\in\R^{2s^2}$ of the filter kernel $k_i^t$. We constrain the filters to be zero-mean which is defined as $\xi_{\reu}^\top \vk_i^t=0,\, \xi_{\imu}^\top \vk_{i}^t=0$, where $\xi_\reu^\top \vk_i^t,\xi_\imu^\top \vk_i^t$ estimate the individual means of the filter kernel on the real and imaginary plane, respectively. Additionally, the whole kernel is constrained to lie on the unit-sphere, i.e.,~$\norm[2]{\vk_i^t}=1$, for simpler parametrization of the activation functions. To learn arbitrary activation functions, we require a suitable parametrization: we define the scalar functions $\phi_i^{t\prime }$ as a weighted combination of $N_w$ Gaussian radial basis functions (RBFs) with equidistant nodes $\mu$ distributed in $[-I_{\textsc{max}},I_{\textsc{max}}]$ and standard deviation $\sigma=\frac{2I_{\textsc{max}}}{N_w - 1}$
	\begin{align*}
		\phi_i^{t\prime }(z) = \sum\limits_{j=1}^{N_w} w_{ij}^t \exp \left( -\frac{(z -
		\mu_j)^2}{2\sigma^2} \right).
	\end{align*}
	Note here that $\mu,\sigma$ depend on the maximum estimated filter response $I_{\textsc{max}}$.
	The final parameters that we consider are the data term weights $\lambda^t$, which are
	constrained to be non-negative ($\lambda^t > 0$). During training, all 	constraints on the parameters are realized based on projected gradient methods.
	
	\subsection{Variational Network Training}
	During the offline training procedure illustrated in Figure~\ref{fig:figure2-training_overview}, the goal is to find an optimal parameter set
	$\theta = \{\theta^0, ... ,\theta^{T-1} \},\, \theta^t = \{w_{ij}^t, \vk_i^t, \lambda^t\}$ for our proposed VN in~\eqref{eq:vn}.
	To set up the training procedure, we minimize a loss function over a set of images $S$ with respect to the parameters $\theta$. The loss function defines the similarity between the reconstructed image $\vu^T$ and a clean, artifact-free reference image $\vg$.
	A common choice for the loss function is the mean-squared error (MSE)
	\begin{align*} \calL(\theta) = \min_\theta \frac{1}{2S}
	\sum_{s=1}^{S}\norm[2]{\vu^{T}_s(\theta) - \vg_{s}}^2.
	\end{align*}
	
	\insertmrmfig{\figII}
	
	As we are dealing with complex numbers in MRI reconstruction and we typically assess
	magnitude images, we define the MSE loss of ($\epsilon$-smoothed) absolute
	values
	\begin{align*}
		\calL(\theta) = \min_\theta \frac{1}{2S} \sum_{s=1}^{S}\norm[2]{\abseps{\vu^{T}_s(\theta)} - \abseps{\vg _{s}}}^2, \quad \abseps{\vx} =\sqrt{\vx _{\textsc{re}}^2 + \vx _{\textsc{im}}^2 + \epsilon}
	\end{align*}
	where $\abseps{\cdot}$ is understood in a point-wise manner.
	To solve this highly non-convex training problem, we use the Inertial Incremental Proximal Gradient (IIPG) optimizer which is related to the Inertial Proximal Alternating Linearized Minimization (IPALM) algorithm~\cite{Pock2016}. For algorithmic details on IIPG refer to Appendix A. First-order optimizers require both the loss function value and the gradient with respect to the parameters $\theta$. This gradient can be computed by simple back-propagation~\cite{LeCun2012}, i.e.,~applying the chain rule
	\begin{align*}
		\fracdd{\calL(\theta)}{\theta^t} = \fracdd{ \vu^{t+1}}{ \theta^t}\cdot\fracdd{ \vu^{t+2}}{
		\vu^{t+1}}\hdots\fracdd{\vu^T}{\vu^{T-1}}\cdot\fracdd{\calL(\theta)}{\vu^T}.
	\end{align*}
	
	The derivation of the gradients for the parameters is provided in Appendix B.
	After training, the parameters $\theta$ are fixed and we can reconstruct previously unseen k-space data efficiently by forward-propagating the k-space data through the VN.
	
	\subsection{Data Acquisition} 		
	A major goal of our work was to explore the generalization potential of a learning based approach for MRI reconstruction. For this purpose, we used a standard clinical knee protocol for data acquisition with a representative patient population that differed in terms of anatomy, pathology, gender, age and body mass index. The protocol consisted of five 2D turbo spin echo (TSE) sequences that differed in terms of contrast, orientation, matrix size and signal-to-noise ratio (SNR). For each sequence,
	we scanned 20 patients on a clinical 3T system (Siemens Magnetom Skyra) using an off-the-shelf
	15-element knee coil. All data were acquired without acceleration, and undersampling was performed retrospectively as needed. The study was
	approved by our institutional review board. Sequence parameters were as follows:\\
	
	\noindent\textbf{Coronal proton-density (PD)}: TR=2750, TE=27ms, TF=4, matrix size $320\times288$, voxel size $0.49\times0.44\times3\text{mm}^3$
	
	\noindent\textbf{Coronal fat-saturated PD}: TR=2870ms, TE=33ms, TF=4, matrix size $320\times288$, voxel size $0.49\times0.44\times3\text{mm}^3$
	
	\noindent\textbf{Axial fat-saturated \Ttwo}: TR=4000ms, TE=65ms, TF=9,
	matrix size $320\times256$, voxel size $0.55\times0.44\times3\text{mm}^3$
	
	\noindent\textbf{Sagittal fat-saturated \Ttwo}: TR=4300ms,
	TE=50ms, TF=11, matrix size $320\times256$, voxel size $0.55\times0.44\times3\text{mm}^3$
	
	\noindent\textbf{Sagittal PD}: TR=2800ms, TE=27ms, TF=4, matrix size $384\times307$, voxel size $0.46\times0.36\times3\text{mm}^3$\\
	
	\noindent Coil sensitivity maps were precomputed from a data block of size $24\times24$ at the center of k-space using
	ESPIRiT~\cite{Uecker2014}. For both training and quantitative evaluation, each
	network reconstruction was compared against a gold standard reference
	image. We defined this gold standard as the coil-sensitivity combined,
	fully sampled reconstruction. The fully sampled
	raw data were retrospectively undersampled for both training and testing.
	
	\subsection{Experimental Setup}
	Our experiments differed in contrast, orientation, acceleration factor and sampling pattern. For all our experiments, we pre-normalized the acquired k-space volumes with $n_\textsc{sl}$ slices by $\frac{\sqrt{n_\textsc{sl}}10000}{\norm[2]{f}}$. We trained an individual VN for each experiment and kept the network architecture fixed for all experiments. The VN consisted of $T=10$ steps. The initial reconstruction $u_0$ was defined by the zero filled solution. In each iteration $N_k=48$ real/imaginary filter pairs of size $11\times 11$ were learned.
	For each of the $N_k$ filters, the corresponding activation function was defined by $N_w=31$ RBFs equally distributed between [-150,150]. Including the data term weight $\lambda^t$ in each step, this resulted in a total of 131,050 network parameters.
	
	For optimization, we used the IIPG optimizer described in Appendix A. The IIPG optimizer allows handling the previously described constraints on the network parameters. We generated a training set for each contrast, sampling pattern and acceleration factor. In each experiment, we used 20 image slices from 10 patients, which amounts to 200 images, as the training set. The training set was split into mini batches of size 10. Optimization was performed for $1000$ epochs with a step size of $\eta=10^{-3}$.
	
	\subsection{Experiments}
	In the first step, we investigated whether the learning-based VN approach actually benefits from structured undersampling artifacts due to regular undersampling, or if it performs better with incoherent undersampling artifacts as are typically present in CS applications. We used a regular sampling scheme with fully-sampled k-space center, identical to the vendor implementation of an accelerated TSE sequence on an MR-system. To introduce randomness, we also generated a variable-density random sampling pattern according to Lustig et al.~\cite{Lustig2007}. Both sampling patterns have the same fully-sampled k-space center and same number of phase encoding steps. We evaluated the acceleration factors $R\in \{3,4\}$ for two sequences which differ in contrast and SNR. The second step was to explore the generalization potential with respect to different contrasts and orientations of a clinical knee protocol.
	
	\subsection{Evaluation}
	We tested our algorithm on data from 10 clinical patients and reconstructed the whole imaged volume for each patient. These cases were not included
	in the training set, and they also contained pathology not represented in the training set. It is worth noting that the number of slices was different for each patient, depending on the individual optimization of the scan protocol by the MR technologist.
	
	We compared our learning-based VN to the linear PI reconstruction method CG SENSE~\cite{Pruessmann1999} and a combined PI-CS non-linear reconstruction method based on Total Generalized Variation (TGV)~\cite{Bredies2010,Knoll2010a}. The forward and adjoint operators for these methods, in particular the coil sensitivity maps, were consistent with our VN approach. All hyper-parameters for CG SENSE and PI-CS TGV such as the number of iterations and regularization parameters were estimated individually by grid search for each sampling pattern, contrast and acceleration factor, such that the MSE of the reconstruction to the gold standard reconstruction was minimized. We assessed the reconstruction results quantitatively in terms of MSE and Structural Similarity Index (SSIM)~\cite{Wang2004} with $\sigma=1.5$ on the magnitude images.

	\subsection{Implementation Details}
	The VN approach as well as the reference methods were implemented in C++/CUDA. We provide Python and Matlab interfaces for testing. Experiments were performed on a system equipped with an Intel Xeon E5-2698 Central Processing Unit (CPU) (2.30GHz) and a single Nvidia Tesla M40 Graphics Processing Unit (GPU). [Note: We will make the source code and data to reproduce the results of the manuscript available when and if the manuscript is accepted for publication.]
	
	\vspace*{1\baselineskip} \section{Results}
	Figures~\ref{fig:figure3-result_cor_pd_r3} and~\ref{fig:figure4-result_cor_pd_r4} display the impact of acceleration factors $R=3$ and $R=4$ and sampling patterns for CG SENSE, PI-CS TGV and our learned VN on coronal PD-weighted images. Additionally, we plot zero filling solutions to illustrate the amount and structure of undersampling artifacts. Residual artifacts and noise amplification can be observed for CG SENSE, in particular for $R=4$. In case of acceleration factor $R=3$, the PI-CS image appears less noisy than CG SENSE; however, similar undersampling artifacts are present. For $R=4$ the PI-CS TGV result contains fewer undersampling artifacts than CG SENSE but the image already appears cartoonish in certain regions. The learned VN suppresses these artifacts while still providing sharper and more natural-looking images. Interestingly, both the PI-CS TGV and learned VN reconstruction with $R=3$ regular sampling perform slightly better than with variable-density random sampling in terms of intensity homogeneity and sharpness. For acceleration $R=4$, randomness improves the reconstruction results. We depict the reconstruction videos of the whole imaged volume for the depicted case and $R=4$ in Supporting Video~\ref{sfig:sfigure1-cor_pd_regular4_video} for regular sampling and in Supporting Video~\ref{sfig:sfigure2-cor_pd_random4_video} for variable-density random sampling. 
	
	Similar observations can be made for coronal PD-weighted scans with fat saturation, as depicted in Figures~\ref{fig:figure5-result_cor_pdfs_r3} and~\ref{fig:figure6-result_cor_pdfs_r4}. The main difference is that this sequence has a lower SNR compared to the non-fat-saturated version. Since additional noise reduces sparsity, the PI-CS TGV reconstructions produce an even more unnatural blocky pattern and contain substantial residual artifacts. Our learned VN is able to suppress these undersampling artifacts and shows improved image quality at this SNR level as well.

	All our observations are supported by the quantitative evaluation depicted in
	Table~\ref{tab:table1-eval}a for regular sampling and in Table~\ref{tab:table1-eval}b for variable-density random sampling. The wide range in quantitative values over the different sequences illustrates the effect of SNR on the reconstructions. The learned VN reconstructions show superior
	performance in terms of MSE and SSIM in all cases. Table~\ref{tab:table1-eval} also supports the qualitative impression that there is no improvement using variable-density random sampling for $R=3$ for PI-CS TGV and VN reconstruction. In contrast, random sampling outperforms regular sampling for $R=4$ in all cases.
	
	\insertmrmfig{\figIII}
	\insertmrmfig{\figIV}
	\insertmrmfig{\figV}
	\insertmrmfig{\figVI}
	
Results for individual scans of a complete knee protocol are illustrated in Figure~\ref{fig:figure7-protocol} along with the zoomed view in Figure~\ref{fig:figure8-protocol_zoom} for regular sampling with $R=4$. These results contain various pathologies, taken from subjects ranging in age from 15 to 57, and anatomical variants, including a pediatric case. In particular, the coronal PD-weighted scan (M50) shows a prior osteochondral allograft transplant indicated by the green arrow. The patient has a history of osteochondritis that was treated with an Osteoarticular Transfer System procedure 18 months prior to the MR. The image shows chondral loss and subchondral bone marrow changes and the patient subsequently underwent an unicompartmental knee arthroplasty.
An extruded and torn medial meniscus, indicated by the green arrow, is visible in the coronal fat-saturated PD-weighted scan. Additionally, this patient (F57) has broad-based, full-thickness chondral loss within the medial compartment and a subchondral cystic change underlying the medial tibial plateau, as indicated by the green bracket. Results for the sagittal PD-weighted scan illustrate a skeletally immature patient (F15) with almost completely fused tibial physes.
A partial tear of the posterior cruciate ligament is visible in the sagittal fat-saturated \Ttwo-weighted scan M34.
A full-thickness chondral defect centered in the medial femoral trochlea (green arrow) is visible on the axial fat-saturated \Ttwo-weighted scan (F45) on a background of patellofemoral osteoarthritis. A reconstruction video of all available image slices for the axial fat-saturated \Ttwo-weighted case is shown in Supporting Video~\ref{sfig:sfigure3-ax_t2fs_ipat4_video}.

\insertmrmfig{\figVII}
\insertmrmfig{\figVIII}

The presence of these particular variations, which were not included in the training data set, does not negatively affect the learned reconstruction. The reduction of residual aliasing artifacts, marked by yellow arrows, the reduced noise level, and the more natural-looking images lead to improved depiction of the pathologies when compared to the reference methods. Again, the quality improvement of the learned VN is supported by the quantitative analysis of similarity measures depicted in Table~\ref{tab:table1-eval}a.

\insertmrmfig{\tabI}

\ifbDraft\clearpage\else\fi\vspace*{1\baselineskip} \section{Discussion}
While deep learning has resulted in clear breakthroughs in Computer Vision, the application of deep learning to medical image reconstruction is just beginning~\cite{Wang2016}. Early attempts to use machine learning for MRI reconstruction were based on dictionary learning~\cite{Ravishankar2011,Caballero2014}. Initial results for our deep learning image reconstruction approach presented in detail here were first presented at the Annual Meeting of the International Society for Magnetic Resonance in Medicine in May of 2016~\cite{Hammernik2016}. Wang et al.~\cite{Wang2016a} showed first results using a convolutional neural network (CNN) architecture to define a relationship between zero filled solution and high-quality images based on pseudo-random sampling. The learned network can then be used as regularization in a non-linear reconstruction algorithm. 
Yang et al.~\cite{Yang2016} introduced a network architecture that is based on
unrolling the Alternating Direction Method of Multipliers algorithm.
They proposed to learn all parameters including image transforms and shrinkage
functions for CS-based MRI. Han et al.~\cite{Han2017} learned destreaking on CT images and then fine-tuned the learning on MR data to remove streaking from radially undersampled k-space data. All three approaches used single-coil data, and it remains unclear how they deal with the complex domain of MR images. Kwon et al.~\cite{Kwon2016} introduced a neural network architecture to estimate the unfolding of multi-coil Cartesian undersampled data. Similar to a classic SENSE reconstruction~\cite{Pruessmann1999}, unfolding is performed line-by-line. This restricts the applicability to a fixed matrix size and a particular 1D undersampling pattern.
Most recently, Lee et al.~\cite{Lee2017} used residual learning to train two CNNs to estimate the magnitude and phase images of Cartesian undersampled data.
	
In this work, we present the first learning-based MRI reconstruction approach for clinical multi-coil data. Our VN architecture combines two fields: variational methods and deep learning. We formulate image reconstruction as a variational model and embed this model in a gradient descent scheme, which forms the specific VN structure.

The VN was first introduced as a trainable reaction-diffusion model~\cite{Chen2015} with application to classic image processing tasks~\cite{Chen2015,Klatzer2016,Yu2015}. All these tasks are similar in the sense that the data are corrupted by unstructured noise in the image domain. MR image reconstruction presents several substantial differences:
complex-valued multi-coil data are acquired in the Fourier domain and transformed into the image domain. This involves the use of coil sensitivity maps and causes distinct artifacts related to the sampling pattern.

One of the key strengths of our proposed VN is the motivation by a generalized, trainable variational model. The solid theoretical foundation of the VN provides insight into the properties of the learned model. This sets it apart from many learning approaches, which are essentially treated as black-boxes where it is very challenging to explain the properties and characteristics of the results. To gain an understanding of what the VN learns, we first inspect the intermediate outputs of the gradient descent steps of our VN (see Supporting Video~\ref{sfig:sfigure4-stage_outputs}). We observe successive low-pass and high-pass filtering, and note that the prevalence of undersampling artifacts decreases after each single iteration. In contrast to iterative reconstruction algorithms, a continuous improvement over the iterations does not occur because our training is designed such that the result after the last gradient step is optimal in terms of the error metric chosen for evaluation. Although it would be possible to train the VN for progressive improvement, this would reduce the flexibility of the algorithm for adjusting the learned parameters during the training procedure. 

In addition, our VN structure allows us to visualize the learned parameters, which is non-trivial for classical CNNs~\cite{Zeiler2014}. The learned filter kernel pairs for real and imaginary feature planes are plotted along with their corresponding activation and potential functions in Figure~\ref{fig:figure9-result_params}. The potential functions are computed by integrating the learned activation functions, and they can be linked directly to the norms that are used in the regularization terms of traditional CS algorithms. Some of the learned filter pairs have the same structure in both the real and imaginary plane while some of them seem to be inverted in the real and imaginary part. In general, the filters in both the real and imaginary part represent different (higher-order) derivative filters of various scales and orientations, similar to Gabor filters~\cite{Gabor1946,Daugman1985}. Handcrafted Gabor filters have been successfully used in image processing~\cite{Jain1990}, and learning-based approaches~\cite{Krizhevsky2012} report similar filters. It has also been shown that these types of filters have a strong relation to the human perceptual system~\cite{Olshausen1996}.

Some of the learned potential functions in Figure~\ref{fig:figure9-result_params} are very close to the convex $l_1$ norm used in CS (e.g.,~the function in the 3rd column), but we can also observe substantial deviations. We can identify functions with student-t characteristics also used in~\cite{Roth2009}, which are reported to fit the statistics of natural images better than, e.g.,~the $l_1$-norm~\cite{Huang1999}. Potential functions like those in columns 1, 6, 9 and 12 have been associated with image sharpening in the literature~\cite{Zhu1997}.

\insertmrmfig{\figIX}

Designing filters and functions is not a trivial task. Using learning-based approaches provides a way to tune these parameters such that they are adapted to specific types of image features and artifact properties. Larger filter sizes, such as the $11\times 11$ filters used in our VN architecture, also provide the possibility to capture more efficiently the characteristic backfolding artifacts of Cartesian undersampled data, which are spread over several pixels. This stands in contrast to models like TV or TGV that are based on gradient filters in a small neighborhood (e.g.,~only forward differences in the x and y direction are considered). To suppress artifacts with PI-CS TGV, the regularization parameters must be chosen in such a way that the remaining image appears over-smoothed, and fine details are lost.
Even though the piecewise-affine prior model of TGV is more complex than the piecewise-constant prior model of TV, the images appear artificial, especially if MR images with low SNR are reconstructed.

In any iterative CS approach, every reconstruction is handled as an individual optimization problem. This is a fundamental difference to our proposed data-driven VN. In our VN approach, we perform the computationally expensive optimization as an offline pre-computation step to learn a set of parameters for a small fixed number of iterations. In our experiments, one training took approximately four days for on a single graphics card. Once the VN is trained, the application to new data is extremely efficient, because no new optimization problem has to be solved and no additional parameters have to be selected. In our experiments, the VN reconstruction took only 193 ms for one slice. In comparison, the reconstruction time for zero filling was 11 ms, for CG SENSE with 6 iterations 75 ms and for PI-CS TGV with 1000 primal-dual iterations~\cite{Knoll2010a} 11.73 s on average. Thus, the online VN reconstruction using the learned parameters for the fixed number of iterations does not affect the hard time constraints during a patient exam.

Our hypothesis based on the CS theory was that all results with a non-linear reconstruction would profit from the randomness introduced with a variable-density random sampling pattern.
When analyzing the reconstruction results, this was surprisingly not the case for a moderate acceleration factor $R=3$. This behavior can be understood as follows: for modest acceleration factors, the gaps in k-space are small enough that the support of the coil-sensitivities, which convolve the underlying k-space data, is sufficient to fill in the missing k-space lines robustly. This is evident in the CG SENSE results for $R=3$, which show almost no residual artifacts (see Figures~\ref{fig:figure3-result_cor_pd_r3} and~\ref{fig:figure5-result_cor_pdfs_r3}). Random sampling increases some of the gaps in k-space, which provides more incoherence at the cost of PI performance. In a situation where the PI component of the reconstruction is already able to remove aliasing, the regularization term in PI-CS TGV mainly acts as a suppressor of g-factor based noise amplification. By contrast, for higher acceleration ($R=4$) both CG SENSE and PI-CS TGV results are strongly corrupted, and in this case randomness leads to an improvement. These results also demonstrate the limits in terms of achieving incoherence with 2D Cartesian sampling. Arguably, the performance of a combined PI-CS method could be improved with a sampling pattern design that provides randomness with an additional constraint on the maximal distance of adjacent lines, such as Poisson disk sampling~\cite{Dunbar2006,Nayak1998,Lustig2010}. The comparison of a wide range of sampling patterns is beyond the scope of this particular manuscript, and will be the target of future work, which will also explore the application of VN reconstruction to non-Cartesian sampling, dynamic and multi-parametric data. Future investigations will also involve the choice of different, e.g.,~perceptual-based, error metrics for training, since MSE and SSIM are likely not optimal for representing similarity to reference reconstructions.

\vspace*{1\baselineskip} \section{Conclusion}
Inspired by variational models and deep learning, we present a new approach, termed VN, for efficient reconstruction of complex multi-coil MR data. We learn the whole reconstruction procedure and all associated model parameters in an offline training step on clinical patient data sets. The VN-based reconstructions preserve important features not presented in the training data. Our proposed learning-based VN reconstruction approach outperforms traditional reconstructions for a wide range of pathologies and offers high reconstruction speed, which is substantial for integration into clinical workflow.

\vspace*{1\baselineskip} \section{Acknowledgements}
We acknowledge grant support from the Austrian Science Fund (FWF) under the START project BIVISION, No. Y729, the European Research Council under the Horizon 2020 program, ERC starting grant ”HOMOVIS”, No. 640156, and from the US National Institutes of Health (NIH P41 EB017183, NIH R01 EB000447), as well as hardware support from Nvidia corporation. We would like to thank Dr. Tobias Block for his support with the Yarra Framework, Dr. Elisabeth Garwood for helping us with clinical evaluation, and Ms. Mary Bruno for assistance with the data acquisition.

\ifbDraft\else\clearpage\fi\section{Appendix A}
\subsection{Inertial Incremental Proximal Gradient Algorithm (IIPG)}
\input{appendix_iipgd}

\section{Appendix B}
\subsection{Gradient Derivation of Network Parameters}
\input{appendix_gradient}

\clearpage \section{References}
\bibliographystyle{mrm}
\bibliography{learning_recon}
	
\ifbDraft
\clearpage \section{Supplementary Material\ifbDraft\footnote{\lowercase{ \url{https://pure.tugraz.at/portal/files/7931056/supplementary_material.zip}}}\else\fi}
\sfigI
\sfigII
\sfigIII
\sfigIV
\losf

\else
\tabI
\sfigI
\sfigII
\sfigIII
\sfigIV

\clearpage \section{List of Figures}
\lof
\clearpage \section{List of Tables}
\lot
\clearpage \section{List of Supporting Videos}
\losf

\clearpage

\figI
\figII
\figIII
\figIV
\figV
\figVI
\figVII
\figVIII
\figIX
\fi
\end{document}

%% file: def_math.tex
\usepackage{mathtools}

\newcommand{\R}{\mathbb{R}}
\newcommand{\C}{\mathbb{C}}

\newcommand{\calR}{\mathcal{R}}
\newcommand{\calL}{\mathcal{L}}

\newcommand{\degree}{\ensuremath{^\circ}\,}

\newcommand{\iu}{\mathrm{j}\mkern1mu}  
\newcommand{\reu}{\textsc{re}}
\newcommand{\imu}{\textsc{im}}

\newcommand{\fracdd}[2]{\frac{\partial #1}{\partial #2}}
\newcommand{\dotp}[2]{\left\langle#1,#2\right\rangle}

\newcommand{\norm}[2][]{\left\|{#2}\right\|_{{#1}}}
\newcommand{\abs}[1]{\lvert #1 \rvert}

\newcommand{\mat}[1]{\ensuremath{\mathbf{#1}}}
\newcommand{\diag}[1]{\operatorname{diag}\left( #1 \right)}  

\newcommand{\Ttwo}{T$_2$}

\newcommand{\abseps}[1]{\vert #1 \vert_\epsilon}
\newcommand{\vu}{\mathbf{u}}
\newcommand{\vk}{\mathbf{k}}
\newcommand{\vg}{\mathbf{g}}
\newcommand{\vf}{\mathbf{f}}
\newcommand{\vz}{\mathbf{z}}
\newcommand{\vx}{\mathbf{x}}
\newcommand{\ve}{\mathbf{e}}
\newcommand{\vw}{\mathbf{w}}

%% file: def_figures.tex
\newcommand{\insertmrmfig}[1]{\ifbDraft #1 \fi}

\newcommand{\figI}{\mrmfigure{\includegraphics[width=0.9\textwidth]{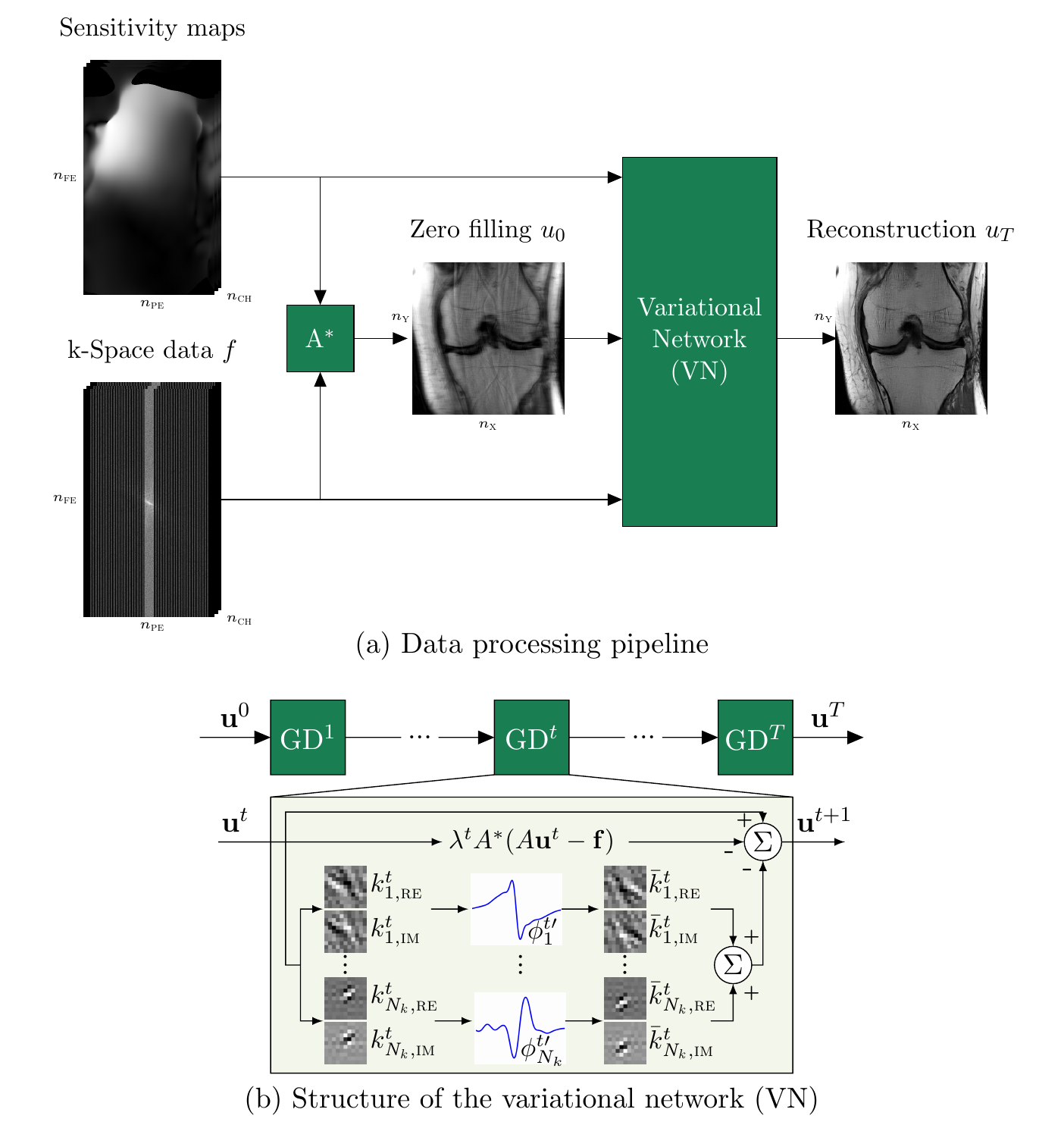}}
{Proposed image reconstruction pipeline and structure of the variational network (VN). (a) A zero filled solution is computed from the undersampled k-space data by applying the adjoint operator $A^*$. The adjoint operator $A^*$ involves application of coil sensitivity maps. We feed the undersampled k-space data, coil sensitivity maps and the zero filling solution to the VN to obtain a reconstruction. For simplicity, we show the magnitude images, but all the input and output data of the VN are complex-valued. The VN consists of $T$ gradient descent steps (b). Here, a sample gradient step is depicted in detail. As we are dealing with complex-valued images, we learn separate filters $k_i^t$ for the real and complex plane. The non-linear activation function $\phi_i^{t\prime}$ combines the filter responses of these two feature planes. During a training procedure, the filter kernels, activation functions and data term weights $\lambda^t$ are learned.}{fig:figure1-network}}

\newcommand{\figII}{\mrmfigure{\includegraphics[width=0.9\textwidth]{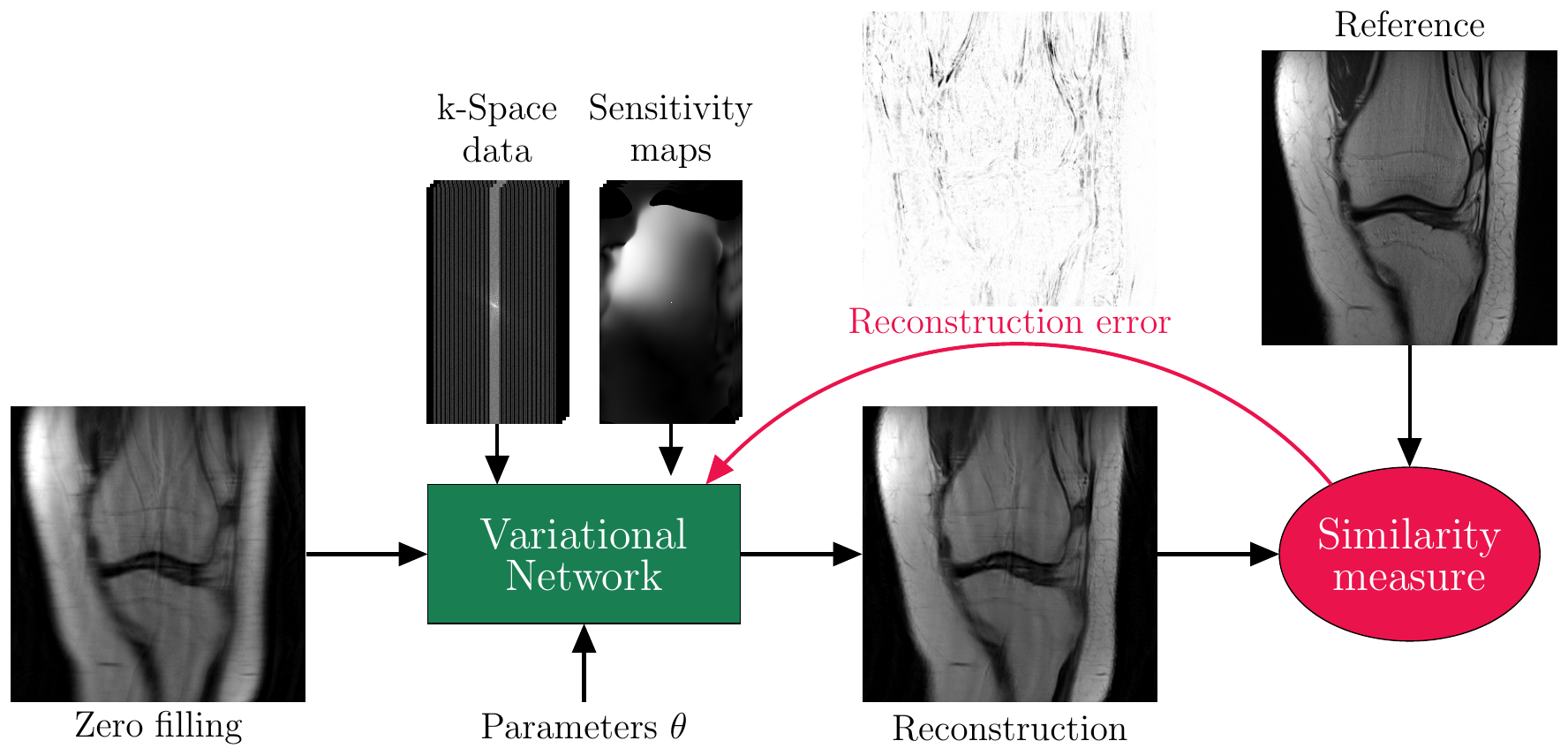}}
{Variational network training procedure: We aim at learning a set of parameters $\theta$ of the VN during an offline training procedure. For this purpose, we compare the current reconstruction of the VN to an artifact-free reference using a similarity measure. This gives us the reconstruction error which is propagated back to the VN to compute a new set of parameters.}{fig:figure2-training_overview}}

\newcommand{\figIII}{\mrmfigure{\includegraphics[width=\textwidth]{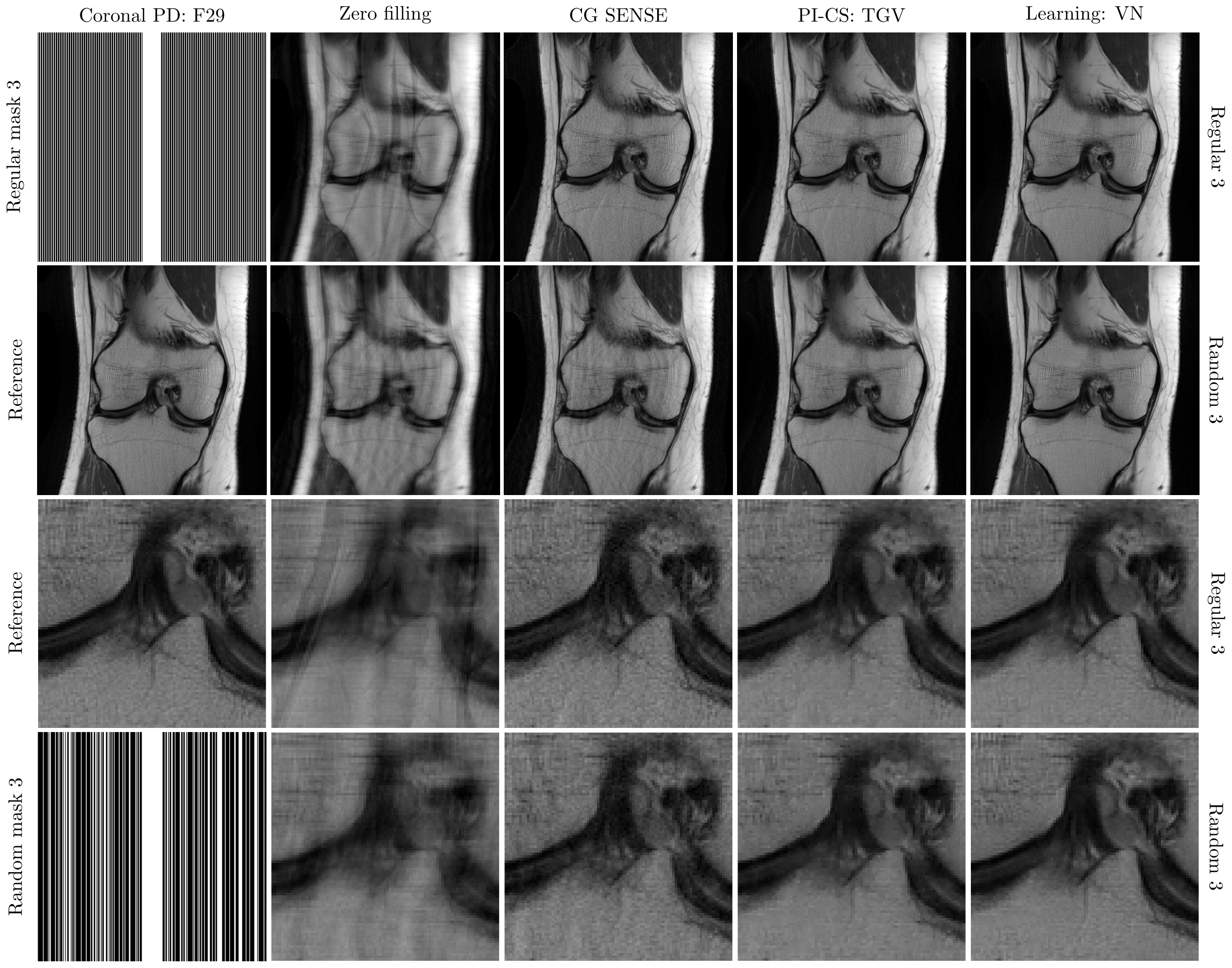}}
{Coronal PD-weighted scan with acceleration $R=3$. The first and third row depict reconstruction results for regular Cartesian sampling, the second and forth row depict the same for variable-density random sampling. Zoomed views show that the learned VN reconstruction appears slightly sharper than the PI-CS TGV reconstruction. For regular sampling, the results illustrate that the VN reconstruction can suppress undersampling artifacts better than CG SENSE and PI-CS TGV. For this acceleration factor of $R=3$, the results based on random sampling appear slightly blurrier than the results based on regular sampling.}{fig:figure3-result_cor_pd_r3}}

\newcommand{\figIV}{
\mrmfigure{\includegraphics[width=\textwidth]{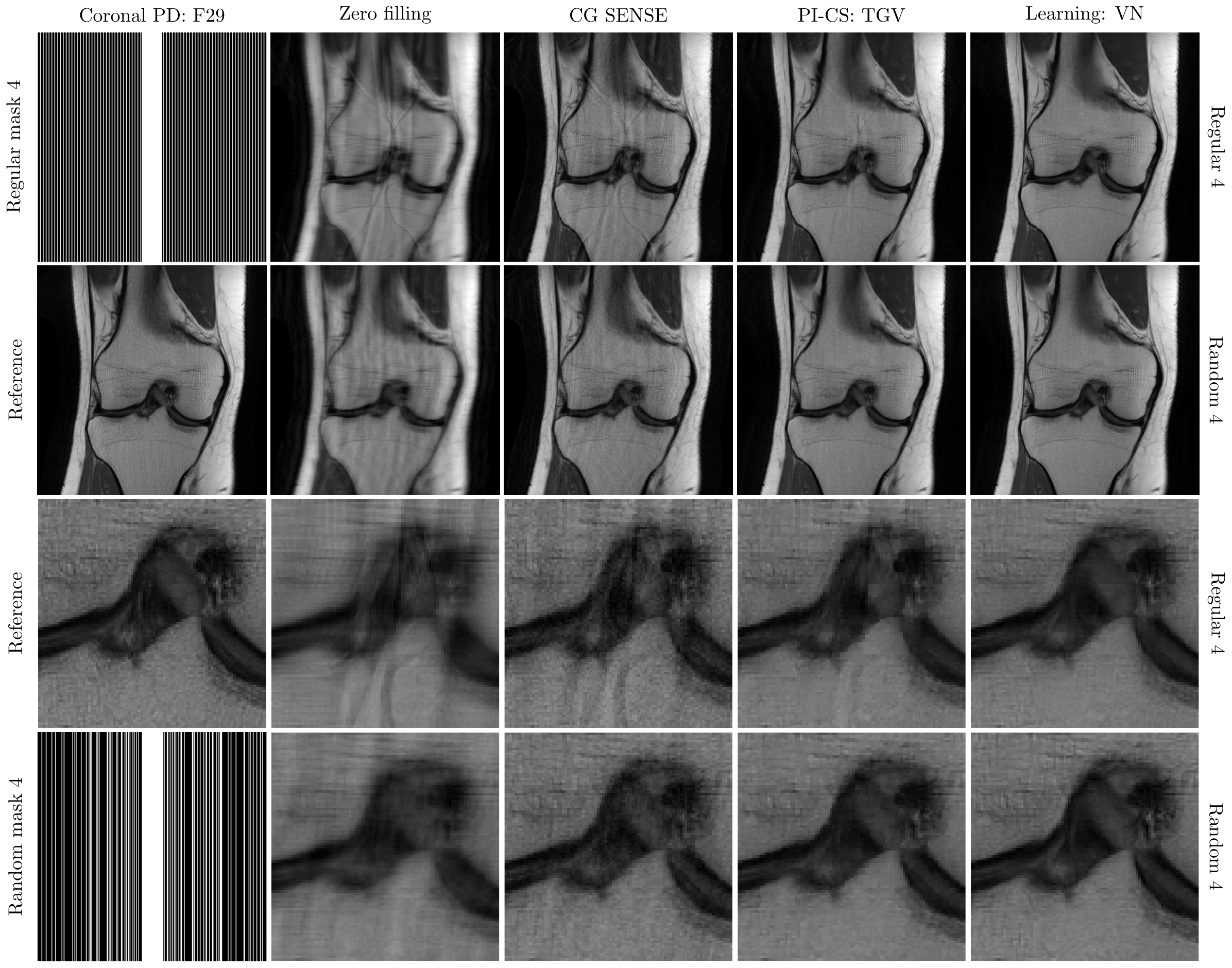}}
{Coronal PD-weighted scan with acceleration $R=4$. The first and third row depict reconstruction results for regular Cartesian sampling, the second and forth row depict the same for variable-density random sampling. Zoomed views show that the learned VN reconstruction appears much more natural than the PI-CS TGV reconstruction. The VN reconstruction can significantly suppress artifacts unlike CG SENSE and PI-CS TGV. Results based on random sampling show reduced residual artifacts and slightly increased sharpness in comparison to regular sampling.}{fig:figure4-result_cor_pd_r4}}

\newcommand{\figV}{\mrmfigure{\includegraphics[width=\textwidth]{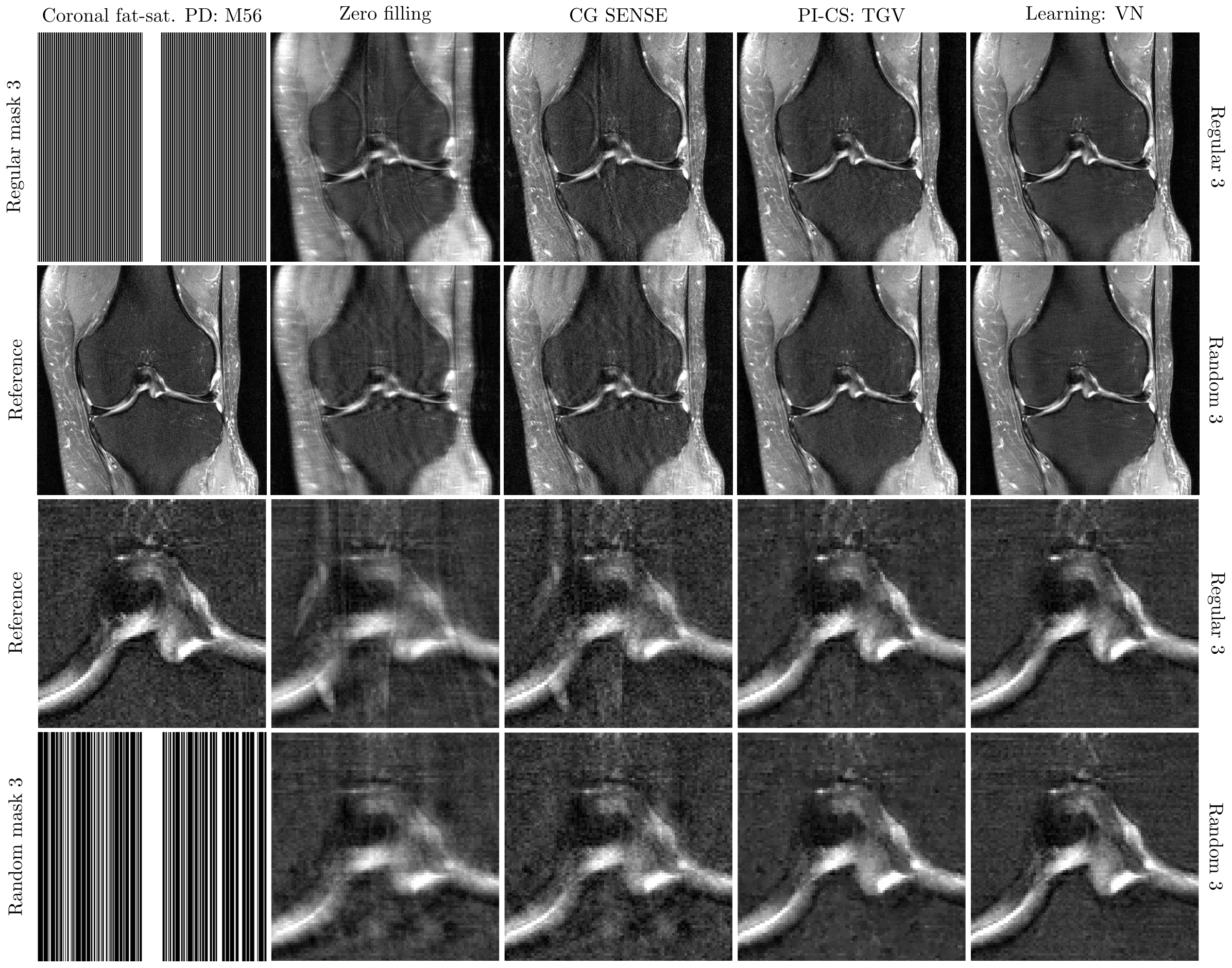}}
{Coronal fat-saturated PD-weighted scan with acceleration $R=3$. The first and third row depict reconstruction results for regular Cartesian sampling, the second and forth row depict the same for variable-density random sampling. The zoomed views show that the learned VN reconstruction appears sharper and more natural than the PI-CS TGV reconstruction. For regular sampling, the results illustrate that the VN reconstruction can suppress undersampling artifacts better. Again, results based on random sampling appear slightly blurrier than the results based on regular sampling.}{fig:figure5-result_cor_pdfs_r3}}

\newcommand{\figVI}{\mrmfigure{\includegraphics[width=\textwidth]{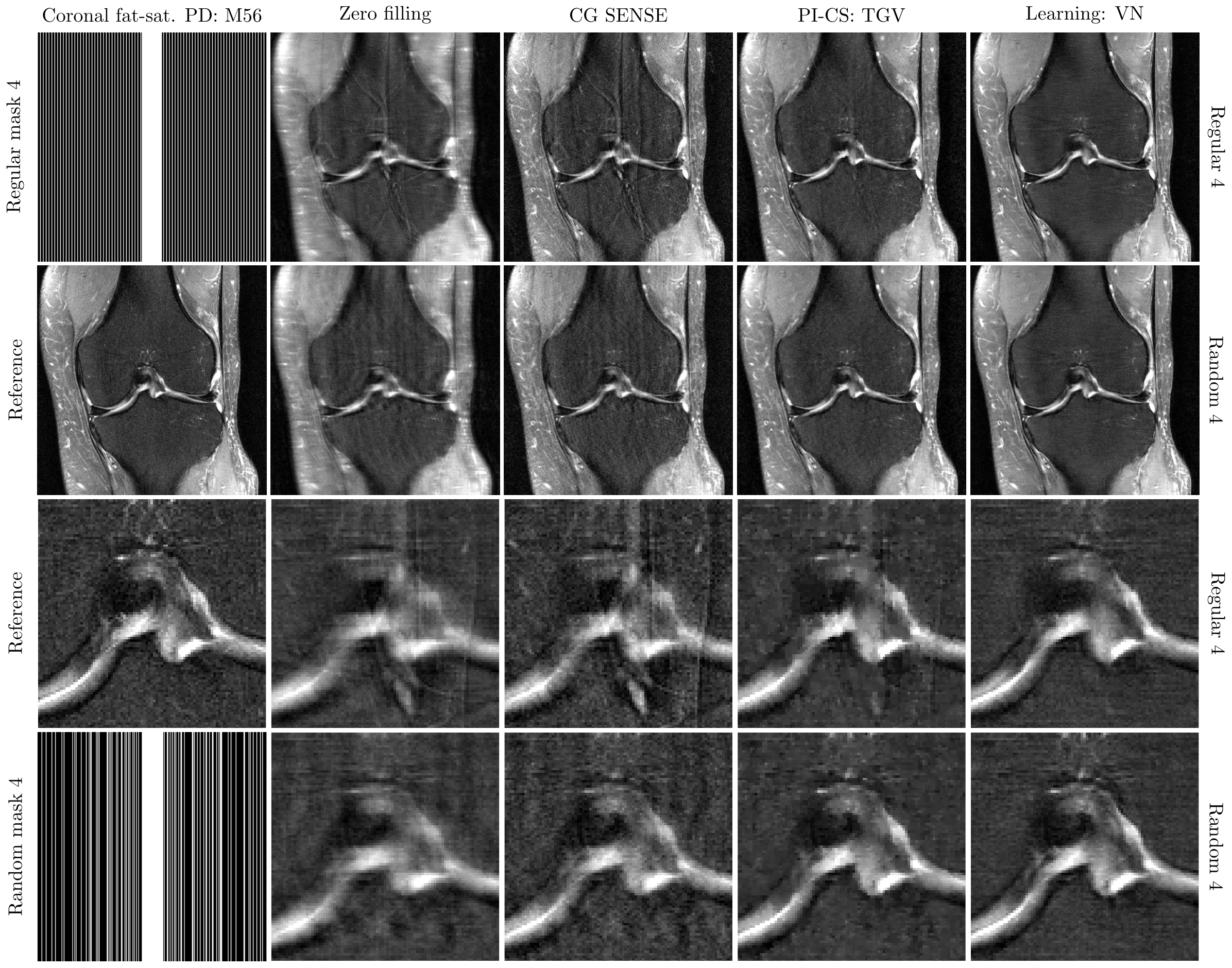}}
{Coronal fat-saturated PD-weighted scan with acceleration $R=4$. The first and third row depict reconstruction results for regular Cartesian sampling, the second and forth row depict the same for variable-density random sampling. The zoomed views show that the learned VN reconstruction appears more natural than the PI-CS TGV reconstruction. The VN reconstruction shows reduced artifacts compared to CG SENSE and PI-CS TGV. Results based on random sampling show reduced residual artifacts and appear sharper than the results based on regular sampling.}{fig:figure6-result_cor_pdfs_r4}}

\newcommand{\figVII}{
\mrmfigure{\includegraphics[width=\textwidth]{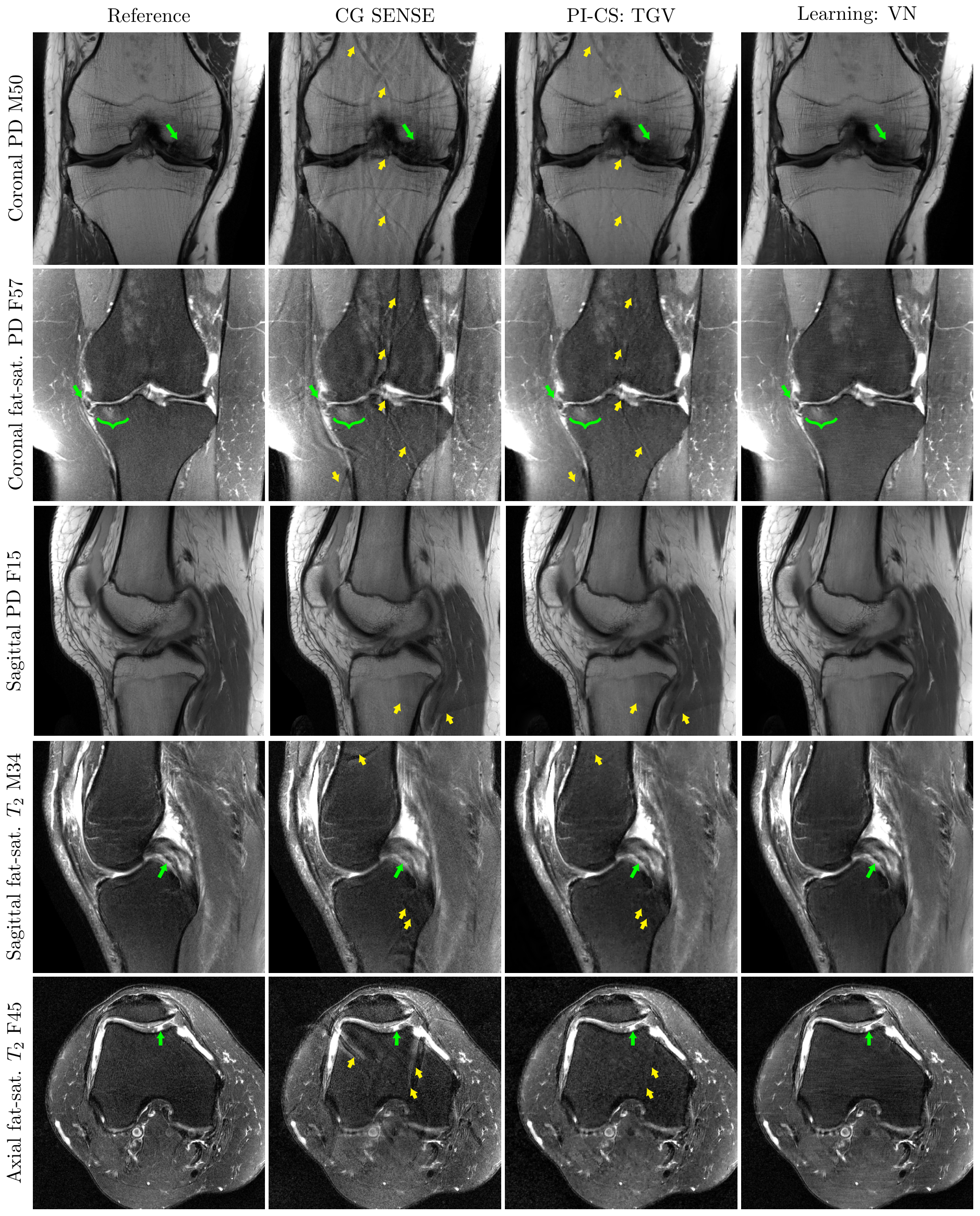}}
{Reconstruction results for a complete knee protocol for acceleration factor $R=4$ with regular undersampling. The protocol includes coronal PD-weighted, coronal fat-saturated PD-weighted, sagittal PD-weighted, fat-saturated sagittal \Ttwo-weighted, and fat-saturated axial \Ttwo-weighted sequences. Each sequence here is illustrated with results from a different patient, identified by gender and age (e.g., M50 indicates a 50-year-old male). Pathological cases and a pediatric case are shown for both male and female patients of various ages. Green arrows and brackets indicate pathologies. Yellow arrows show residual artifacts that are visible in the CG SENSE and PI-CS TGV reconstructions, but not in the learned VN reconstructions.}{fig:figure7-protocol}}

\newcommand{\figVIII}{\mrmfigure{\includegraphics[width=\textwidth]{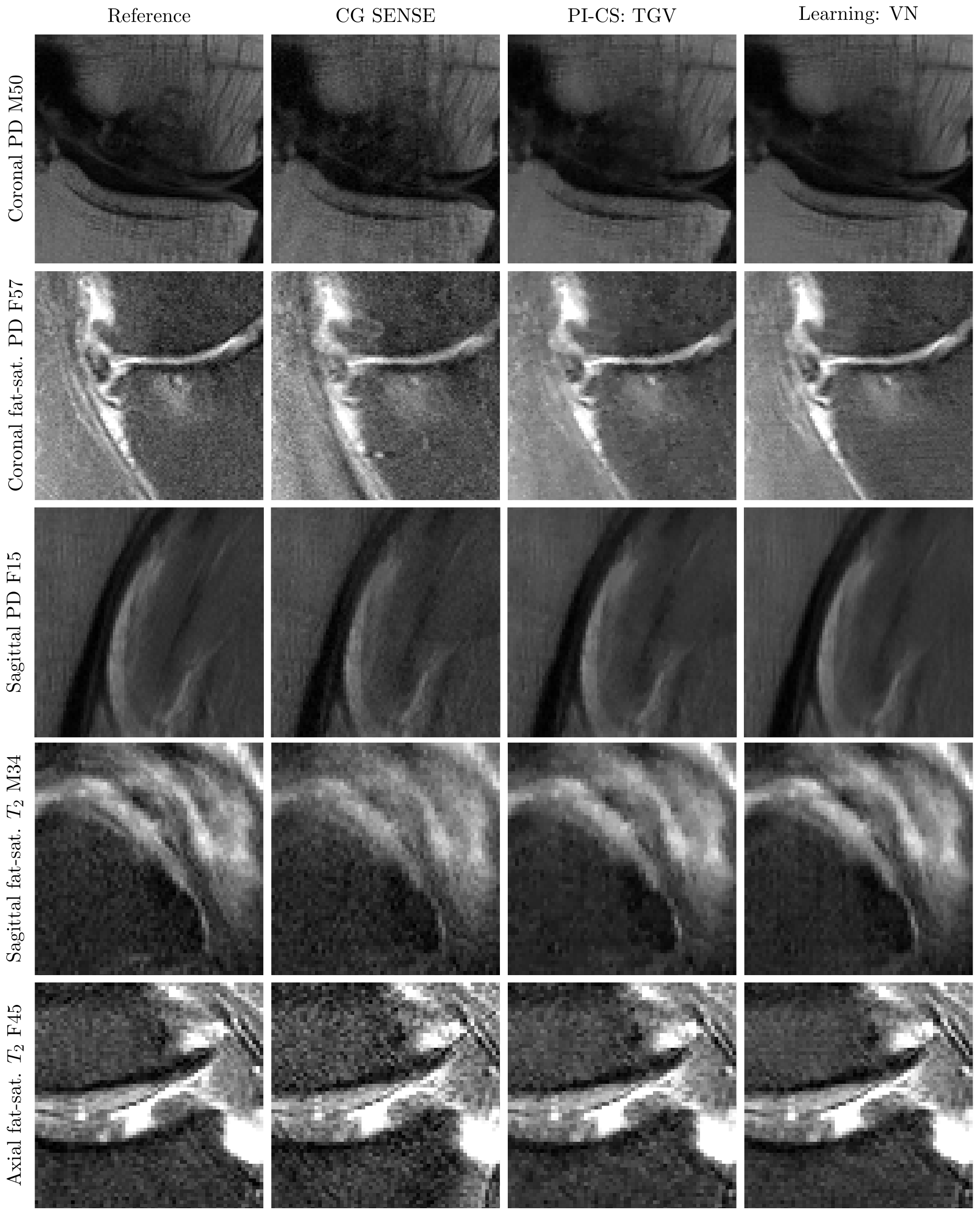}}	{Detailed view of reconstruction results from Fig.~\ref{fig:figure7-protocol} for a complete clinical knee protocol for $R=4$.}{fig:figure8-protocol_zoom}}

\newcommand{\figIX}{\mrmfigure{\includegraphics[width=\textwidth]{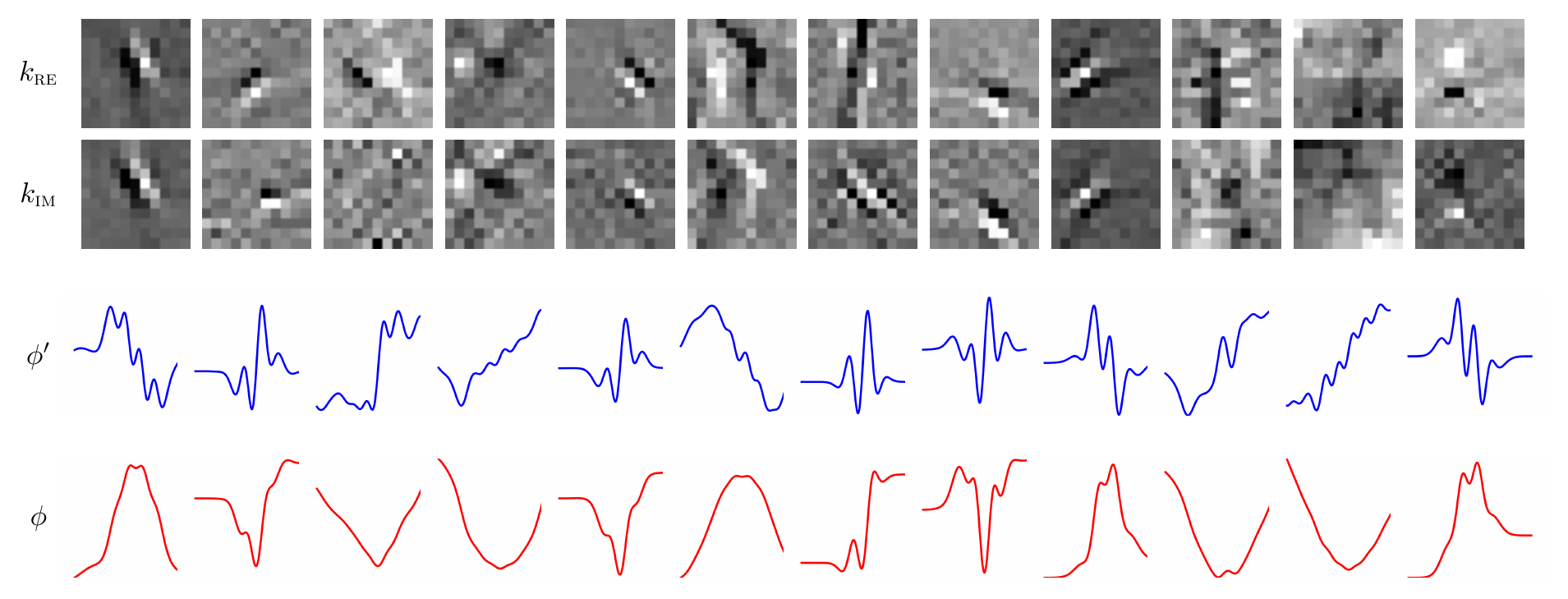}}{Examples of learned parameters of the VN. Filter kernels for the real $k_{\textsc{re}}$ and imaginary $k_{\textsc{im}}$ plane as well as their corresponding activation $\phi'$ and potential function $\phi$ are shown.}{fig:figure9-result_params}}

\newcommand{\tabI}{\mrmtable{
		\includegraphics[width=\textwidth]{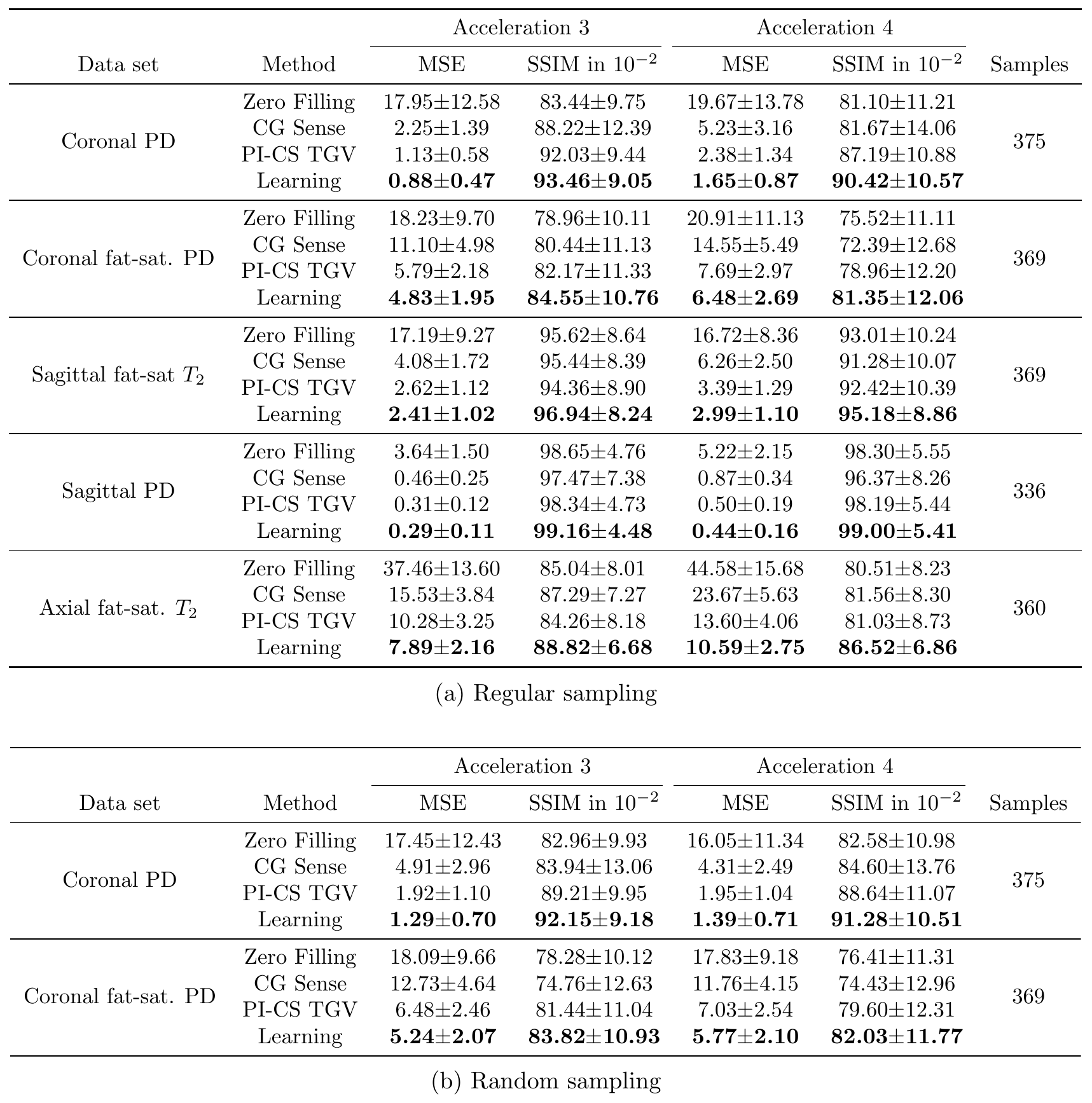}
		} {Quantitative evaluation results in terms of MSE and SSIM for five clinical knee data sets and acceleration factors $R\in\{3,4\}$. (a) shows the quantitative results for regular sampling, and (b) shows the corresponding results for variable-density random sampling.}{tab:table1-eval}}

\newcommand{\sfigI}{\mrmsfigure{%
}
{Reconstruction of a complete imaged volume for a coronal PD-weighted sequence in a 50-year-old male, for regular sampling with acceleration $R=4$.}{sfig:sfigure1-cor_pd_regular4_video}}

\newcommand{\sfigII}{\mrmsfigure{%
}
{Reconstruction of a complete imaged volume for a coronal PD-weighted sequence in the same 50-year-old male patient as in Supporting Figure 1, for variable-density random sampling with acceleration $R=4$.}{sfig:sfigure2-cor_pd_random4_video}}

\newcommand{\sfigIII}{\mrmsfigure{%
}
{Reconstruction of a complete imaged volume for an axial fat-saturated \Ttwo-weighted sequence in a 45-year-old female patient, for regular sampling with acceleration $R=4$.}
{sfig:sfigure3-ax_t2fs_ipat4_video}}

\newcommand{\sfigIV}{\mrmsfigure{%
}%
{%
	Intermediate gradient step outputs of the reconstruction algorithm for a coronal PD-weighted slice with acceleration $R=4$. We observe alternating low-pass and high-pass filtering over the intermediate steps. The undersampling artifacts are continuously suppressed until we obtain an artifact-free image after the final step.
}{sfig:sfigure4-stage_outputs}}

%% file: appendix_iipgd.tex
For network training, we consider following optimization problem:
\begin{gather*}
\calL(\theta) = \min_\theta \frac{1}{2S} \sum_{s=1}^{S}\norm[2]{\abseps{\vu^{T}_s(\theta)} - \abseps{\vg_{s}}}^2 \quad \theta = \{\theta^0, ... ,\theta^{T-1} \},\, \theta^t = \{w_{ij}^t, \vk_i^t, \lambda^t\}\\
\vu^{t+1}_s = \vu^t_s - \sum_{i=1}^{N_k}(K_i^t)^\top \Phi_i^{t\prime}(K_i^t \vu^t_s) - \lambda^t A^*(A\vu^t_s - \vf_s), \quad 0 \leq t \leq T-1\\
\text{s.t.} \quad \theta \in \mathcal{C} = \left\lbrace \lambda^t \geq 0, \, \xi_\reu^\top \vk_i^t = 0,\, \xi_\imu^\top \vk_i^t = 0,\, \norm[2]{\vk_i^t} = 1 \right\rbrace.
\end{gather*}

To solve this highly non-convex training problem, we use the Inertial Incremental Proximal Gradient (IIPG) optimizer. This IIPG variant of projected gradient descent is related to the Inertial Proximal Alternating Linearized Minimization (IPALM) algorithm~\cite{Pock2016}. In the IIPG Algorithm~\ref{algo:iipgd}, the parameter updates are calculated on a single mini batch. First, we perform over-relaxation where we set a over-relaxation constant $\beta_e$ dependent on the current epoch $e$ to achieve moderate acceleration. Second, we compute the gradient with respect to the parameters on the current mini batch which yields a new parameter update $\tilde{\theta}^{m+1}$ for the current iteration $m$. To realize additional constraints on the parameters, we finally perform the projections
\begin{align*}
\left(\lambda^{m+1},\,\vk^{m+1}\right) = \text{proj}^{\eta }_{\mathcal{C}}\left(\tilde{\lambda}^{m+1},\,\tilde{\vk}^{m+1}\right).
\end{align*}
As the constraints do not depend on each other, we can consider the projections independently. To realize the non-negativity constraint on the data term weights $\lambda^{m+1}$, the parameter update $\tilde{\lambda}^{m+1}$ is clamped at zero
\begin{align*}
\lambda^{m+1} = \max (0, \tilde{\lambda}^{m+1}).
\end{align*}
For the projection onto the filter kernel constraints, we first subtract the means $ \xi_\reu^\top \tilde{\vk}^{m+1},\, \xi_\imu^\top \tilde{\vk}^{m+1}$ from the current kernel parameter estimates and then project the kernel onto the unit-sphere
\begin{gather*}
\tilde{\vk}^{m+1}_\xi = (\tilde{\vk}^{m+1}_{\xi,\reu},\tilde{\vk}^{m+1}_{\xi,\imu}) = (\tilde{\vk}^{m+1}_\reu - \xi_\reu^\top \tilde{\vk}^{m+1},\tilde{\vk}^{m+1}_\imu - \xi_\imu^\top \tilde{\vk}^{m+1})\\
\vk^{m+1} = \frac{\tilde{\vk}^{m+1}_\xi}{\norm[2]{\tilde{\vk}^{m+1}_\xi}}.
\end{gather*}

\begin{algorithm}[t]
	\textbf{Input:} Split training set $\mathcal{S}$ into $N_B$ mini batches $\mathcal{B}$  s.t.  $\mathcal{S}=\bigcup_{b=1}^{N_B}\mathcal{B}_b$\;
    \textbf{Choose:} Step size $\eta$, number of epochs $N_E$, initial parameters $\theta^0$\;
	Iteration $m \leftarrow 1$, $\theta^1 \leftarrow \theta^0$\;
	\For{$e\leftarrow 1$ \KwTo $N_E$}
	{
		// Over-relaxation constant\\
		$\beta_e=\frac{e-1}{e+2}$\;
		\For{$b\leftarrow 1$ \KwTo $N_B$}
		{
			// Over-relaxation\\
			$\hat{\theta}^{m+1} = \theta^m + \beta_e (\theta^m - \theta^{m-1})$\;
			// Compute gradient on current mini batch  $\mathcal{B}_b$\\
			$g^{m+1} = \frac{ \partial \calL(\hat{\theta}^{m+1})}{\partial\theta}$\;
			// Compute gradient step\\
			$\tilde{\theta}^{m+1} = \hat{\theta}^{m+1} - \eta g^{m+1}$\;
			// Compute projections\\
			$\theta^{m+1} = \text{proj}^{\eta }_{\mathcal{C}}(\tilde{\theta}^{m+1})$\;
			$m \leftarrow m + 1$\;
		}
	}
\caption{Inertial Incremental Proximal Gradient (IIPG) Algorithm}\label{algo:iipgd}
\end{algorithm}

%% file: appendix_gradient.tex
In every gradient step $t$, we seek the derivatives with respect to the parameters $\theta^t=\{w_{ij},\vk_i^t,\lambda ^t\}$ of the loss function
\begin{align*}
\calL(\theta) = \min_\theta \frac{1}{2S} \sum_{s=1}^{S}\norm[2]{\abseps{\vu^{T}_s(\theta)} - \abseps{\vg_{s}}}^2, \quad \abseps{\vx} =\sqrt{\vx_\textsc{re}^2 + \vx_\textsc{im}^2 + \epsilon}
\end{align*}
where $\abseps{\cdot}$ is understood in a point-wise manner. For simplicity, we drop the dependency of $\vu^T$ on the parameters $\theta$ and the subscript $s$ and show the calculations only for a single training example.
The gradient steps are given as
\begin{align*}
\vu^{t+1} = \vu^t - \sum_{i=1}^{N_k}(K_i^t)^\top \Phi_i^{t\prime}(K_i^t \vu^t) - \lambda^t A^*(A\vu^t - \vf), \quad 0 \leq t \leq T-1.
\end{align*}

The derivatives with respect to the parameters $\theta^t$ are obtained by back-propagation~\cite{LeCun2012}
\begin{align*}
\fracdd{\calL(\theta)}{\theta^t} =
\fracdd{\vu^{t+1}}{\theta^t}\cdot\underbrace{\fracdd{\vu^{t+2}}{
	\vu^{t+1}}\hdots\fracdd{
	\vu^T}{\vu^{T-1}}\cdot\fracdd{\calL(\theta)}{\vu^T}}_{\ve^{t+1}}.
\end{align*}
The reconstruction error of the $t$-th gradient step is given by $\fracdd{\calL(\theta)}{\vu^{t+1}} = \ve^{t+1}$.

\paragraph{Derivative of the Loss Function}
First, we require the gradient of the loss function $\calL$ with respect to the reconstruction $\vu^T$ defined as $\ve^T$. It is computed as
\begin{align*}
\frac{\partial\calL(\theta)}{\partial \vu^T} = \ve^T \Leftrightarrow e_l^T=\frac{u_l^T}{\abseps{u_l^T}}\left(\abseps{u_l^T} - \abseps{g_l}\right),\quad l=1,...,N.
\end{align*}

\paragraph{Derivative of the Data Term Weights $\lambda^t$}
The derivative of the reconstruction $\vu^t$ wrt. to $\lambda^t \in \R$ for the $t$-th gradient step is expressed as:
\begin{align*}
\fracdd{\calL(\theta)}{\lambda^t} = \fracdd{\vu^{t+1}}{\lambda^t} \fracdd{\calL(\theta)}{ \vu^{t+1}} = \dotp{-(A^*(A\vu^t - \vf))}{\ve^{t+1}}.
\end{align*}

\paragraph{Derivative of the Activation Functions $\Phi_i^{t\prime}$}
A single activation function $\Phi_i^{t\prime}(\vz) = \left(\phi_i^{t\prime}(z_1) ,...,\phi_i^{t\prime}(z_N)  \right): \R^N \mapsto \R^N$ is defined by a weighted combination of $N_w$ Gaussian radial basis functions:
\begin{align*}
\phi_i^{t\prime}(z_l) = \sum\limits_{j=1}^{N_w}w_{ij}^t \exp{\left(-\frac{(z_l-\mu_j)^2}{2\sigma^2}\right)},\quad l=1,...,N,\quad w_{ij}^t \in \R.
\end{align*}
This can be rewritten in a matrix-vector notation:
\begin{align*}
\newcommand{\rbf}[2]{\exp{\left(-\frac{(z_{#1}-\mu_{#2})^2}{2\sigma^2}\right)}}
\Phi_i^{t\prime}(z)=
\begin{pmatrix}
\phi_i^{t\prime}(z_1) \\
\vdots\\
\phi_i^{t\prime}(z_N)
\end{pmatrix}
=
\begin{bmatrix}
\rbf{1}{1} & \hdots & \rbf{1}{N_w} \\
\vdots & \ddots & \vdots\\
\rbf{N}{1} & \hdots & \rbf{N}{N_w}
\end{bmatrix}
\begin{pmatrix}
w_{i1}^t \\ \vdots \\ w_{iN_w}^t
\end{pmatrix}
=\mat{M}_i^t(\vz)\vw_i^t.
\end{align*}
During training, we learn the weights $\vw_i^t \in \R^{N_w}$ and express its gradient as:
\begin{align*}
\fracdd{\calL(\theta)}{\vw_i^t} = \fracdd{\vu^{t+1}}{\vw_i^t}\fracdd{\calL(\theta)}{\vu^{t+1}}  = -\fracdd{}{ \vw_i^t}\left\lbrace(K_i^t)^\top  \mat{M}_i^t(K_i^t \vu^t)\vw_i^t \right\rbrace \ve^{t+1} = -\left(\mat{M}_i^t(K_i^t \vu^t)\right)^\top K_i^t \ve^{t+1}.
\end{align*}

\paragraph{Derivative of the Intermediate Reconstructions $\vu^t$}
Further gradients with respect to the reconstructions from intermediate steps are given as:
\begin{align*}
\fracdd{\vu^{t+1}}{\vu^{t}} = I - \sum_{i=1}^{N_k}(K_i^t)^\top \diag{\Phi_i^{t\prime\prime}(K_i^t \vu^t)}K_i^t - \lambda^t A^*A
\end{align*}
where $I$ denotes the identity matrix. This also requires the second derivative of the potential functions $\Phi_i^{t\prime\prime}(\vz)$, which is expressed as:
\begin{align*}
\newcommand{\rbf}[2]{\exp{\left(-\frac{(z_{#1}-\mu_{#2})^2}{2\sigma^2}\right)}}
\newcommand{\drbf}[2]{-\frac{(z_{#1}-\mu_{#2})}{\sigma^2}}
\Phi_i^{t\prime\prime}(\vz)=
\begin{bmatrix}
\drbf{1}{1}\rbf{1}{1} & \hdots & \drbf{1}{N_w}\rbf{1}{N_w} \\
\vdots & \ddots & \vdots\\
\drbf{N}{1}\rbf{N}{1} & \hdots & \drbf{N}{N_w}\rbf{N}{N_w}
\end{bmatrix}
\vw_i^t
\end{align*}

\paragraph{Derivative of the Filter Kernels $k_i^t$}
To compute the derivative with respect to the filter kernels $\ k_i^t$ we have to introduce further relationships between our given parameters. The convolution can be defined as matrix-vector multiplication:
\begin{align*}
k_i^t \ast u^t \Leftrightarrow K_i^t \vu^t = U^t\vk_i^t
\end{align*}
where the matrix $U^t:\R^{2s^2}\mapsto\R^{N}$ is a suitably shifted representation of the image $\vu^t$ and $\vk_i^t\in\R^{2s^2}$ is the vectorized filter kernel.
The gradient step also involves rotated filter kernels $\bar{k}_i^t$ due to the transpose operation of the kernel matrix $(K_i^t)^\top$. As we want to calculate the derivative with respect to $\vk_i^t$ and not to their rotated version, we introduce a rotation matrix $R: \R^{2s^2} \mapsto \R^{2s^2}$ that has the same effect as the transpose operation
\begin{align*}
\bar{\vk}_i^t=R\vk_i^t.
\end{align*}
The convolution can be rewritten as
\begin{align*}
(K_i^t)^\top \Phi_i^{t\prime}(K_i^t\vu^t) = \tilde{\Phi}^{t\prime}_i(K_i^t\vu^t)\bar{\vk}_i^t = \tilde{\Phi}^{t\prime}_i(K_i^t\vu^t)R\vk_i^t
\end{align*}
where $\tilde{\Phi}_i^t(K_i^t\vu^t):\R^{N}\mapsto\R^{2s^2}$ is a suitable matrix representation of $\Phi_i^t(K_i^t\vu^t)$.
Applying the product rule yields following expression for the kernel derivative
\begin{align*}
\fracdd{(K_i^t)^\top \Phi_i^{t\prime}(K_i^t \vu^t)}{\vk_i^t} = \fracdd{\Phi_i^{t\prime}(K_i^t \vu^t)}{\vk_i^t}K_i^t +
\fracdd{\vk_i^t}{\vk_i^t}\left[\tilde{\Phi}^{t\prime}_i(K_i^t\vu^t) R\right]^\top = \\
(U^t)^\top\diag{\Phi_i^{t \prime \prime}(K_i^t\vu^t)}K_i^t + R^\top\tilde{\Phi}_i^{t\prime}(K_i^t \vu^t).
\end{align*}
The full derivative may be expressed as
\begin{align*}
	\fracdd{\calL(\theta)}{ \vk_i^t} = \fracdd{ \vu^{t+1}}{ \vk_i^t}\fracdd{\calL(\theta)}{ \vu^{t+1}} = - \left[(U^t)^\top\diag{\Phi_i^{t \prime \prime}(K_i^t\vu^t)}K_i^t + R^\top\tilde{\Phi}_i^{t\prime}(K_i^t \vu^t)\right] \ve^{t+1}.
\end{align*}